\documentclass[11pt,a4paper]{article}
\usepackage{times,latexsym,framed}
\usepackage{url}
\usepackage[T1]{fontenc}

\usepackage[acceptedWithA]{tacl2021v1}

\usepackage{microtype}
\usepackage{booktabs}
\usepackage{multirow}
\usepackage{enumitem,kantlipsum}
\usepackage{siunitx}
\usepackage{chngpage}
\usepackage{amssymb}
\usepackage{changes}
\usepackage{mathtools}
\usepackage{enumitem}
\usepackage{lipsum}
\usepackage{dblfloatfix}

\newcommand{\result}[2]{ #1 \color{lightgray}\!{\scriptsize{$\pm{#2}$}}}
\newcommand{\icc}[3]{{\scriptsize\textcolor{lightgray}{#1}} #2 {\scriptsize\textcolor{lightgray}{#3}}}

\newcommand{\minkprob}{\textsc{Min-K\% Prob}}

\newcommand{\ignore}[1]{}

\title{Do Language Models Enjoy Their Own Stories?\\Prompting Large Language Models for Automatic Story Evaluation}

\author{
    Cyril Chhun$^1$ \qquad Fabian M. Suchanek$^1$ \qquad Chloé Clavel$^2$ \\
    $^1$LTCI, Télécom Paris, Institut Polytechnique de Paris \\
    $^2$ALMAnaCH, INRIA Paris\\
    \texttt{cyril.chhun@telecom-paris.fr}\\
    \texttt{fabian.suchanek@telecom-paris.fr}\\
    \texttt{chloe.clavel@inria.fr}\\
}

\date{}

\begin{document}
\maketitle

\begin{abstract}
  Storytelling is an integral part of human experience and plays a crucial role in social interactions. Thus, Automatic Story Evaluation (ASE) and Generation (ASG) could benefit society in multiple ways, but they are challenging tasks which require high-level human abilities such as creativity, reasoning and deep understanding. Meanwhile, Large Language Models (LLM) now achieve state-of-the-art performance on many NLP tasks. In this paper, we study whether LLMs can be used as substitutes for human annotators for ASE. We perform an extensive analysis of the correlations between LLM ratings, other automatic measures, and human annotations, and we explore the influence of prompting on the results and the explainability of LLM behaviour. Most notably, we find that LLMs outperform current automatic measures for system-level evaluation but still struggle at providing satisfactory explanations for their answers.
\end{abstract}

\section{Introduction}

The task of Automatic Story Generation (ASG) \citep{li2013story} consists in the creation of a narrative from a short sentence. Previous research showed that storytelling enables a narrator to communicate honestly with their audience \citep{rowcliffe2004storytelling} and to provide listeners with an engaging and instructive experience \citep{miller2008power}. Indeed, the process of story creation is a salient testimony of human creativity, requiring both the discovery of interesting ideas and their adept expression through a carefully-built narrative. Strong automatic story generating systems could therefore be useful for a variety of applications, such as gaming \citep{turner2014creative}, education \citep{lombardo2012storytelling}, mental health \citep{george2014creative} and marketing \citep{junior2023story}.

Meanwhile, over the last few years, advances in natural language processing (NLP) have been spearheaded by the development of large language models (LLM) such as GPT-3 \citep{brown2020language}, LaMDA \citep{thoppilan2022lamda}, PaLM \citep{chowdhery2022palm} and LLaMA \citep{touvron2023llama}. Upon release, these models have been setting new state-of-the-art performance standards for a wide array of NLP tasks, \textit{e.g.}\ question answering, summarization, and translation. In particular, for ASG, LLMs are now able to produce convincing stories, so much so that they can be hard to distinguish from human stories \citep{clark2021all}. As their performance improves, they may become valuable assistants to our creative process; already, writing contests have been shown to encourage their use \citep{edilivre2023concours}.

The increased availability of LLMs to the general public underlines the need for reliable story evaluation methods that can be used to improve both the performance of ASG models and our understanding of their strengths and weaknesses. Since the human annotation of stories is costly and time-consuming \citep{celikyilmaz2020evaluation}, Automatic Story Evaluation (ASE) systems could provide an efficient and scalable replacement for human evaluation. However, current automatic measures have been shown to be poorly correlated with human judgment for ASG \citep{chhun-etal-2022-human}. 

\begin{figure*}[!t]
    \centering
    \includegraphics[width=\textwidth]{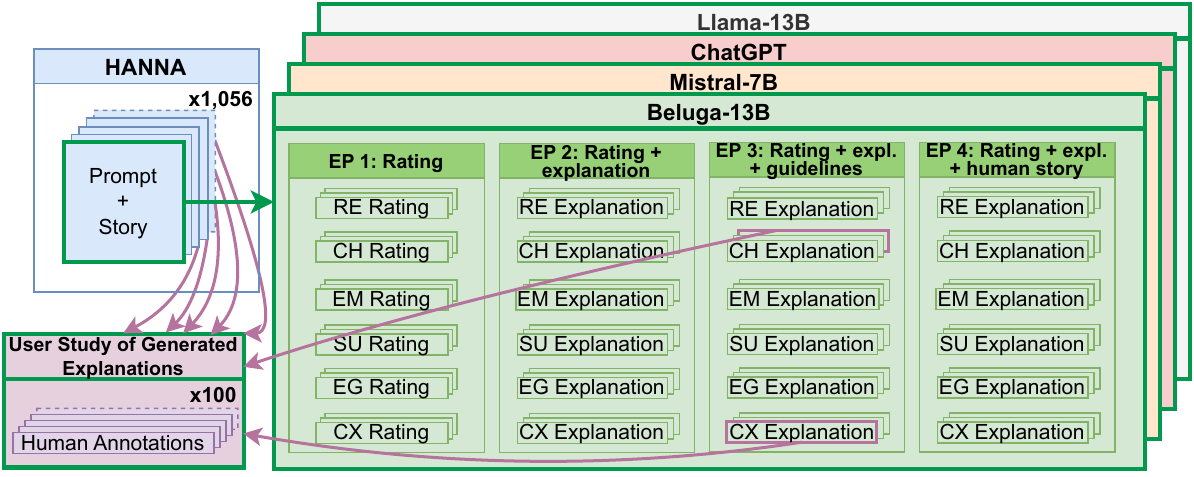}
    \caption{Schema of the performed ASE experiments. RE, CH, etc. are the considered human criteria (\autoref{sub:methodology}). ``EP'' means ``Eval-Prompt'', defined in \autoref{sub:methodology}. For the user study (\autoref{ssub:user_study}), we randomly sampled 100 explanations from our experiments.}
    \label{fig:llm_schema}
\end{figure*}

In this paper, we investigate whether LLMs themselves can be used as substitutes for human annotators for story evaluation. 
To that end, we perform several annotation experiments where we ask different LLMs to rate stories according to different criteria and to explain their rating. We show an example in \autoref{fig:demo_example} and a schema of our experiments in \autoref{fig:llm_schema}.

Our contributions are the following:

\begin{enumerate}[wide, labelindent=0pt]
    \item \textbf{A comparison between LLMs and current ASE measures.} We compute and analyze the correlations between LLM ratings with human annotations on criteria specific to story evaluation; we find that, while overall correlations are moderate to weak, system-level correlations are very high, suggesting that LLMs can produce reliable model rankings for ASE;
    \item \textbf{An analysis of the influence of prompt engineering on LLM performance.} We examine the effects of using different Eval-Prompts on the consistency and distribution of LLM ratings. We find that adding detailed guidelines does not necessarily improve performance and that LLMs are remarkably self-consistent;
    \item \textbf{Insights on LLM explainability for ASE.} We analyze the explanations provided by LLMs through different methods, including a user study, and find that LLMs' understanding of the ASE task is perfectible. Most notably, they struggle at explaining their ratings with substantiated claims;
    \item \textbf{An analysis of LLM performance in ASG.} The high system-level correlations of LLMs with human ratings enable us to use them to rate other LLMs for ASG. We find that LLMs perform at least as well as humans for the generation of short stories, and that their performance may be explained by their tendency to produce output that is similar to their pretraining data.
\end{enumerate}

\begin{figure}[t!]
    \small
    \begin{framed}
    \underline{Eval-Prompt}\\
    \noindent Prompt: A mirror shows your reflection and your future soulmate. You only see your reflection.\\
    \noindent Story: Carnival music seemed to dance in the background. A cacophony of color and knick knacks decorated the fairgrounds. [...]\\
    Rate the story on a scale from 1 to 5 on Empathy and explain your answer.
    \end{framed}
    \begin{framed}
    \underline{LLM Answer}\\
    I would rate the story a 2 on Empathy. While I could relate to the character's feelings of indifference towards the mirror, I did not feel a strong emotional connection to any of the characters. [...]
    \end{framed}
    \caption{Example Eval-Prompt and answer from our experiments. ``Prompt'' inside the Eval-Prompt refers to the story-prompt.}
    \label{fig:demo_example}
\end{figure}

Our methodology can be found in \autoref{sub:methodology}. We release our data and code on GitHub\footnote{\url{https://github.com/dig-team/hanna-benchmark-asg}}. Our data consists of:
\begin{itemize}[nolistsep]
    \item \textbf{ASE experiments}: $\sim$150k rating and explanation annotations (1,056 stories, 6 criteria, 4 Eval-Prompts, 3 tries, 2 models);
    \item \textbf{User study}: 1,500 human annotations of LLM explanations;
    \item \textbf{ASG experiment}: 384 stories generated by Llama models with corresponding LLM annotations to expand the \texttt{HANNA} dataset of \citet{chhun-etal-2022-human}.
\end{itemize}

\noindent This paper is structured as follows: in \autoref{sec:related_work}, we review the related work. In \autoref{sec:llms_for_asg}, we lay out our methodology and experimental details. In \autoref{sec:analysis}, we perform our analysis of the results. In \autoref{sec:discussion}, we discuss the state of LLMs in ASG and ASE. Finally, in \autoref{sec:conclusions}, we conclude with practical takeaways for researchers, the limitations of our work, and future research directions.

\section{Related work}
\label{sec:related_work}

\subsection{Human Evaluation}
Evaluating stories is a difficult task \citep{mccabe1984makes, dickman2003four}. In the social sciences literature, multiple criteria have been suggested, often divided into cognitive and emotional factors \citep{bae2021preliminary}. However, the consensus around the criteria to be used in the NLP literature is still weak \citep{fan2018hierarchical, guan-etal-2020-knowledge, rashkin-etal-2020-plotmachines, goldfarb-tarrant-etal-2020-content}. \citet{chhun-etal-2022-human} distill the indicators used in the social sciences literature into 6 criteria (Relevance, Coherence, Empathy, Surprise, Engagement, Complexity), which we will use in our paper as well.

While human evaluation remains the gold standard of evaluation, it is costly and time-consuming. We therefore need to develop automatic measures that can act as substitutes for human judgment, ideally for each of the criteria. Such automatic measures could be used to improve language models, \textit{e.g.}\ as a loss function or for chain-of-thought prompting \citep{wei2022chain}.

\subsection{Automatic Evaluation}
Automatic measures (\textit{e.g.}\ BLEU \citep{papineni2002bleu}, ROUGE \citep{lin2004rouge}, BERTScore \citep{zhang2019bertscore}, BARTScore \citep{yuan2021bartscore}) have been repeatedly shown to correlate moderately to poorly with human judgment, especially when applied to tasks other than the one they were designed for \citep{zhang2004interpreting, novikova2017we, colombo2022glass}. \citet{deutsch2022limitations} put forth the particular limitations of reference-free measures. For ASE, \citet{guan-etal-2020-knowledge} and \citet{chhun-etal-2022-human} also observe weak correlations between automatic and human ratings, whether they be reference-based or reference-free. This highlights the need for better automatic evaluation methods.
To tackle this issue, this paper investigates the use of LLMs to annotate stories with ratings w.r.t.\ a given criterion.

\subsection{Automatic Annotation}
LLMs are increasingly being tested for automatic text annotation, \textit{e.g.}\ for sentiment analysis  \citep{qureshi2022novel}, named entity recognition \citep{enkhsaikhan2021auto} or event structure modeling \citep{vauth2021automated}. \citet{wang2021want} demonstrate that labeling performed by GPT-3 can achieve the same performance as human labeling and be up to 96\% more cost-efficient. \citet{ding2022gpt} show that GPT-3 performs well for text classification tasks, but struggles with more complex tasks such as named entity recognition. \citet{chakrabarty2023art} design a test for creativity and show that LLM-generated stories pass fewer tests than human stories, and that using LLMs for ASE yields no positive correlations.

We seek to generalize their findings through the use of source-available models and a finer analysis and discussion of LLM performance.

\subsection{Prompt Engineering}
The importance of designing efficient prompts for large language models such as GPT-3 has been extensively investigated in recent years. \citet{reynolds2021prompt} notably find that zero-shot prompting can perform similarly to few-shot prompting, and even exceed it. They explore the design of metaprompts that prime the language model to better solve a given problem. \citet{zhou2022large} treat the prompt engineering process as an optimization problem, use search algorithms guided by LLMs to solve it and attain human-level performance. \citet{wei2022emergent} and \citet{white2023prompt} review different strategies that have been applied to augment large language model abilities, \textit{e.g.}\ least-to-most prompting \citep{zhou2022least}, ask-me-anything prompting \citep{arora2022askma}, and zero-shot chain-of-thought reasoning \citep{kojima2022large}.

We choose to investigate whether LLMs perform better with simple or detailed guidelines, and with zero- or one-shot Eval-Prompts.

\begin{figure*}[h!]
    \small\centering
    \begin{minipage}{0.25\textwidth}
        \begin{framed}
        \underline{Eval-Prompt 1}\\
        \\
        \noindent \textbf{Prompt}: You have become death, destroyer of worlds.\\
        \\
        \noindent \textcolor{black}{\textbf{Target Story}: You look up to see all of them in fear. You just must fix this soon. Slowly, just like your Father always had instructed him, you look down and see all your foes dead and beaten down. You can't resist the urge to touch the wounds. For there is nothing you can do about it. [...]}\\
        \\
        Rate the story on a scale from 1 to 5 on Surprise (how surprising the end of the story was). Rating:
        \end{framed}
    \end{minipage}
    \hspace{0.1cm}
    \begin{minipage}{0.45\textwidth}
        \begin{framed}
        \underline{Eval-Prompt 3}\\
        \\
        \noindent \textbf{Prompt}: You have become death, destroyer of worlds.\\
        \\
        \noindent \textbf{Target Story}: You look up to see all [...]\\
        \\
        \noindent \textbf{Guidelines}:\\
        1 — The ending seemed completely obvious from the start, or doesn’t make any sense at all.\\
        2 — The ending was easily predictable after a few sentences.\\
        3 — The ending was predictable after half of the story.\\
        4 — The ending surprised you, but would have been difficult to predict.\\
        5 — The ending surprised you, and still seemed as if it could very reasonably have been predicted, ie, there were enough clues in the story.\\
        \\
        Rate the story on a scale from 1 to 5 on Surprise (how surprising the end of the story was) and explain your answer. Use the provided guidelines. Rating:
        \end{framed}
    \end{minipage}
    \hspace{0.1cm}
    \begin{minipage}{0.25\textwidth}
        \begin{framed}
        \underline{Eval-Prompt 4}\\
        \\
        \noindent \textbf{Prompt}: You have become death, destroyer of worlds.\\
        \\
        \noindent \textbf{Target Story}: You look up to see all [...]\\
        \\
        \noindent \textbf{Human Story}: I saw the button. It was simple, red, no words on it as I already knew what it did. I mean I built the button, I built what happens [...]\\
        \\
        Rate the target story on a scale from 1 to 5 on Surprise (how surprising the end of the story was) and explain your answer. Do not rate the human story; it is here only for reference. Rating:
        \end{framed}
    \end{minipage}
    \caption{Example Eval-Prompts for the Surprise criterion. Eval-Prompt 2 is the same as Eval-Prompt 1 with ``explain your answer'' added at the end. ``Prompt'' (bold) refers to the story-prompt.}
    \label{fig:gpt3_prompts}
\end{figure*}

\section{Meta-Evaluation of LLMs for ASE}
\label{sec:llms_for_asg}

\subsection{Methodology for ASE}
\label{sub:methodology}

The ASG task commonly involves the generation of a story from a short sentence called a \textit{prompt} \citep{alabdulkarim2021automatic}, which we will henceforth call \textit{story-prompt}.

\paragraph{ASE Definition.}
Given an evaluation measure $m$ (\textit{e.g.}\ a scoring algorithm, an LLM\dots), a story-prompt $i$, and a story $y_i$, we define the ASE task as the production of an evaluation score $m(y_i)$.

In this paper, we choose to use LLMs as ASE measures. We will refer to the prompt that is fed to the LLM as the \textit{Eval-Prompt}, to distinguish it from the story-prompt. See \autoref{fig:demo_example} for an example of the use of an LLM for story evaluation.

\paragraph{ASE Criteria.} We use the criteria introduced by \citet{chhun-etal-2022-human}, who designed \texttt{HANNA}, a benchmark for story evaluation. They compiled a set of six orthogonal criteria from the social sciences literature:
\begin{enumerate}[nolistsep]
    \item \textbf{Relevance} (\texttt{RE}, how well the story matches its prompt), 
    \item \textbf{Coherence} (\texttt{CH}, how much the story makes sense), 
    \item \textbf{Empathy} (\texttt{EM}, how well the reader understood the character’s emotions),
    \item \textbf{Surprise} (\texttt{SU}, how surprising the end of the story was),
    \item \textbf{Engagement} (\texttt{EG}, how much the reader engaged with the story),
    \item \textbf{Complexity} (\texttt{CX}, how elaborate the story is).
\end{enumerate}

\paragraph{Methodology.}
Given the importance of good prompt engineering \citep{zhao2021calibrate}, we design four different Eval-Prompts for the generation of ratings. For each of our Eval-Prompts, we provide the model with a story-prompt and a corresponding story. Then:

\textbf{Eval-Prompt 1} (simple rating): we ask the model to rate the story on a scale from 1 to 5 on one of the six criteria;

\textbf{Eval-Prompt 2} (rating with explanation): same as Eval-Prompt 1, and we ask the model to explain its answer;

\textbf{Eval-Prompt 3} (rating with explanation and guidelines): same as Eval-Prompt 2, and we provide the model with the detailed guidelines from the original annotation protocol by \citet{chhun-etal-2022-human};

\textbf{Eval-Prompt 4} (rating with explanation and human story): same as Eval-Prompt 2, and we provide the model with the human story associated with the same story-prompt. We explicitly tell the model that the human story is given only for reference purposes.

Different Eval-Prompt examples are shown in \autoref{fig:gpt3_prompts}.

\subsection{Meta-Evaluation Measures}
\label{ssec:math_background}
\textbf{Notations.} For $S$ systems and $N$ story-prompts, let $y_i^j$ be the story generated by system $j \in \{1,\dots,S\}$ for story-prompt $i \in \{1,\dots,N\}$. For a (human or automatic) measure $m$, we denote by $m(y_i^j)$ the score associated to $y_i^j$. Let $K$ be a correlation coefficient, \textit{e.g.}\ Pearson's $r$ \citep{pearson1895vii}, Spearman's $\rho$ \citep{spearman1961proof} or Kendall's $\tau$ \citep{kendall1938new}. We note $h_k$ the measure provided by the $k$-th human annotator.

A naive method to compare ratings from two measures would be to compute how much they differ from each other for each story, \textit{e.g.}\ by calculating the average L1 distance between a given evaluation method $m$ and the human ratings, \textit{i.e.}, $\frac{1}{3} \sum_{k=1}^3 \mathcal{L}_1 (m, h_k)$. However, this method suffers from the central tendency bias---the tendency of an individual to rate most items on a survey in the middle of a rating scale---which is often observed in Likert scales \citep{stevens1971issues} and could be explained by the participants' tendency to base their judgment on a least mean squares estimator rather than a maximum a posteriori estimator \citep{douven2018bayesian}. We therefore choose more robust measures of meta-evaluation: system-level and overall correlations.

\paragraph{System-level correlation ($K^\textrm{sys}_{m_1,m_2}$).} We take the correlation of the vectors containing the mean score of all stories for each system, for $m_1$ and $m_2$. This strategy measures how much $m_1$ and $m_2$ agree when comparing different systems. Formally:
\begin{align}
K^\textrm{sys}_{m_1,m_2} &\triangleq K \left(\frac{1}{N} \mathbf{C}^\textrm{sys}_{m_1},\frac{1}{N} \mathbf{C}^\textrm{sys}_{m_2} \right),\\
\text{where} \quad \mathbf{C}^\textrm{sys}_{m} & \triangleq \left[ \sum\limits_{i=1}^N m(y_i^1),\dots, \sum\limits_{i=1}^N m(y_i^S)\right].\nonumber
\end{align}
\vspace*{-0.5\baselineskip}

The segment-level correlation, often used in conjunction with the system-level one in the meta-evaluation literature \citep{ma-etal-2019-results,bhandari2020re}, is not adapted to ASE since stories generated from the same story-prompt are not required to be similar, while \textit{e.g.}\ translations of a sentence should look alike. We therefore use the overall correlation, which we define below.

\paragraph{Overall Correlation ($K_{m_1,m_2}$).} We take the correlation between the full vectors containing the scores of $m_1$ or $m_2$ for a given story for every system. Formally:
\begin{align}
K_{m_1,m_2} & \triangleq K (\mathbf{C}_{m_1},\mathbf{C}_{m_2}),\\
\text{where} \quad \mathbf{C}_{m} & \triangleq \left[\left(m(y_i^j)\right)_{(i,j) \in \{1, \dots, N\} \times \{1, \dots, S\}}\right].\nonumber
\end{align}
\vspace*{-0.5\baselineskip}

\paragraph{Statistical Testing (\autoref{sub:ase1_analysis}).} Correlations between two automatic measures on the same annotated dataset are not independent. As advised by \citet{graham-baldwin-2014-testing}, we use the Williams test \citep{williams1959regression,moon2019williams} to evaluate the strength of an increase in dependent correlations \citep{steiger1980tests}.

Given three features $X_1$, $X_2$ and $X_3$ of a population of size $n$, Williams's $t$ test for whether the correlation between $X_1$ and $X_2$ equals the correlation between $X_1$ and $X_3$ is formulated as follows:
\[ t = \frac{(r_{12} - r_{13}) \sqrt{(n-1)(1+r_{23})}}{\sqrt{2K \frac{(n-1)}{(n-3)} + \frac{(r_{12} + r_{13})^2}{4} (1 - r_{23})^3}}, \]
where $r_{ij}$ is the correlation between $X_i$ and $X_j$ and
\[ K = 1 - {r_{12}}^2 - {r_{13}}^2 - {r_{23}}^2 + 2 \, r_{12} \, r_{13} \, r_{23} . \]
Williams's $t$ statistic follows a Student's $t$-distribution with $n-3$ degrees of freedom. In particular, the Williams test 
takes the correlations between $X_2$ and $X_3$ into account.

Furthermore, since we perform a large quantity of tests, we choose to correct $p$-values for multiplicity. As advised by \citet{jafari2019and}, we control the false discovery rate using the Benjamini-Hochberg (BH) method \citep{benjamini1995controlling}
. Given $n$ $p$-values $p_1, \dots, p_n$ sorted in increasing order, the BH method consists in 
computing adjusted $p$-values $p_k^\star = p_k \frac{m}{k}$ and replacing the $p$-values from largest to smallest.

Following recent recommendations to move beyond simplistic ``statistical significance'' tests \citep{amrhein2019scientists, wasserstein2019moving, mcshane2019abandon}, we report all $p$-values for transparency. 
We choose to use a gradual notion of evidence for our statistical analysis, as suggested by \citet{muff2022rewriting}.

\subsection{Human Evaluation of ASE Explanations}
\label{ssub:user_study}
We conduct a user study in which we ask human raters to identify potential issues in LLM explanations. \citet{dou-etal-2022-gpt} introduced an error annotation schema called \textsc{Scarecrow} that we adapted for ASE. We manually reviewed a random sample of 20 explanations from Beluga-13B on Eval-Prompt~3 and selected the most relevant error types. Then, we randomly sampled another 100 explanations and, for each explanation, we asked 3 human workers to annotate it w.r.t.\ the following five error categories:
\begin{enumerate}[nolistsep]
    \item \textbf{Poor Syntax}: parts of the explanation are grammatically incorrect or wrongly-worded;
    \item \textbf{Incoherence}: parts of the explanation are self-contradictory, logically wrong, or simply do not make sense and do not fit the other categories;
    \item \textbf{Wrong Guideline}: the explanation does not respect the provided guidelines;
    \item \textbf{Superfluous Text}: parts of the explanation contain text that repeats itself or generation artefacts;
    \item \textbf{Unsubstantiated Claims}: the explanation fails to make explicit references to the story to substantiate its reasoning.
\end{enumerate}
We recruited workers on Amazon Mechanical Turk. We estimated that a HIT would
take around one minute, so we set the
reward at \$0.20 per HIT, so about \$12 per hour. To ensure that annotators
spoke fluent English, we restricted access to the
experiment to the UK, the US,
Canada, Australia and New Zealand. 

\subsection{Experimental Details}
\paragraph{Dataset.}
We use the \texttt{HANNA} dataset \citep{chhun-etal-2022-human} which contains 1,056 stories generated from story-prompts from the \texttt{WritingPrompts} dataset \citep{fan2018hierarchical}, with both pretrained language models: \textbf{BERTGeneration} \citep{rothe2020leveraging}, \textbf{CTRL} \citep{keskar2019ctrl}, \textbf{GPT} \citep{radford2019language}, \textbf{GPT-2} \citep{radford2019language}, \textbf{RoBERTa} \citep{liu2019roberta} and \textbf{XLNet} \citep{yang2019xlnet}; and ASG-specific models: \textbf{Fusion} \citep{fan2018hierarchical}, \textbf{HINT} \citep{guan2021long} and \textbf{TD-VAE} \citep{wilmot2021temporal}. These stories were annotated with scores from human raters on the six criteria introduced in \autoref{sub:methodology} and 72 automatic measures. We reproduce the original procedure from \citet{chhun-etal-2022-human}: for reference-based evaluation measures (\textit{e.g.}\ \texttt{BLEU}), we use the human story from \texttt{HANNA} as the reference for the generated story. Because of space constraints, we display only the evaluation measures that are the most used in the literature: \texttt{BLEU} \citep{papineni2002bleu}, \texttt{ROUGE} \citep{lin2004rouge}, \texttt{chrF} \citep{popovic2015chrf}, \texttt{BERTScore} \citep{zhang2019bertscore}, \texttt{SUPERT} \citep{gao2020supert}, \texttt{BLANC} \citep{vasilyev2020fill}, \texttt{BARTScore} \citep{yuan2021bartscore}, \texttt{BaryScore} \citep{colombo2021automatic}. The results are similar for the other automatic measures.

\paragraph{ASG Models.}
Since the release of the \texttt{HANNA} dataset, language models have made significant advancements. We therefore felt the need to expand \texttt{HANNA} with more recent models. We selected \textbf{Llama-2-7b-chat-hf} (Llama-7B) as a new baseline and 4 high-performing models (at the time of selection) of different sizes on the HuggingFace Open LLM Leaderboard\footnote{\url{https://huggingface.co/spaces/HuggingFaceH4/open_llm_leaderboard}}: \textbf{Platypus2-70B-instruct} (Platypus2), \textbf{Llama-30b-instruct-2048} (Llama-30B), \textbf{StableBeluga-13B} (Beluga-13B), \textbf{Mistral-7B-OpenOrca} (Mistral).

\paragraph{ASE Models.}
We submit each of the four Eval-Prompts 3 times on all 1,056 stories on each of the 6 criteria, and we then extract the ratings automatically from the generated answer via a regular expression. Since story evaluation on multiple prompts and multiple criteria was more computationally demanding, we limited our experiments to the smaller 13B and 7B models. We used the 4 following models: Beluga-13B, Mistral, \textbf{Llama-2-13b-chat-hf} (Llama-13B), and \textbf{Gpt-3.5-turbo} (ChatGPT). We also ran the ASE experiments with Llama-7B, which failed at the task too often for the results to be exploitable, \textit{e.g.}\ by generating nonsensical conversations between itself and the user. We use $(\textrm{temperature}, \textrm{top\_p}) = (1, 0.95)$ for Llama models and $(0.7, 1)$ for ChatGPT (default suggested values).

Llama2 \citep{touvron2023llama2} models were trained on a closed ``new mix of data from publicly available sources''. Beluga-13B and Mistral-7B are Llama2 models fine-tuned on Orca-style datasets which contain triplets of ``System message--User
query--LLM response'' for a large collection of tasks \citep{mukherjee2023orca}. Beluga-13B is fine-tuned on StabilityAI's closed internal dataset, while Mistral-7B is fine-tuned on the open OpenOrca dataset \citep{OpenOrca}. ChatGPT \cite{brown2020language, ouyang2022training} is a closed-source model trained on a closed internal dataset that includes the CommonCrawl, Books1 and Books2 datasets.

We used the \textbf{transformers} library \citep{wolf-etal-2020-transformers} and the OpenAI API for our experiments.

\section{Analysis of the results}
\label{sec:analysis}
Our work aims at answering five important questions for ASE and ASG:
\begin{itemize}[wide, labelindent=0pt]
    \item \textbf{ASE1}: How do LLMs compare w.r.t.\ current evaluation methods, both human and automatic?
    \item \textbf{ASE2}: How does the Eval-Prompt influence the consistency and distribution of LLM ratings?
    \item \textbf{ASE3}: How explainable is the evaluation performed by LLMs?
    \item \textbf{ASG1}: Relying on ASE results, how do LLMs perform at ASG?
    \item \textbf{ASG2}: How does pretraining data help predict ASG performance?
\end{itemize}

\subsection{ASE1: Comparison with Current Evaluation Measures}
\label{sub:ase1_analysis}

\subsubsection{Automatic Annotation Consistency}
\label{ssub:icc1}

\begin{table}[!h]
\centering
\begin{tabular}{cccc}
\toprule
Crit. & Beluga-13B & Mistral-7B & Human \\
\midrule
\texttt{RE} & \result{0.88}{0.01} & \result{0.86}{0.01} & \result{0.48}{0.30} \\
\texttt{CH} & \result{0.93}{0.01} & \result{0.90}{0.01} & \result{0.29}{0.28} \\
\texttt{EM} & \result{0.88}{0.01} & \result{0.87}{0.02} & \result{0.34}{0.09}\\
\texttt{SU} & \result{0.80}{0.02} & \result{0.63}{0.03} & \result{0.28}{0.12}\\
\texttt{EG} & \result{0.91}{0.01} & \result{0.87}{0.01} & \result{0.46}{0.12}\\
\texttt{CX} & \result{0.85}{0.01} & \result{0.78}{0.02} & \result{0.56}{0.08}\\
\bottomrule
\end{tabular}
\caption{Intra-class coefficients type 2k for Eval-Prompt 1 ratings with 95\% confidence interval. Higher is better.}
\label{tab:icc1}
\end{table}

First, we want to verify if LLMs provide stable answers. The default decoding strategy for LLMs (both Llama models and ChatGPT) is top-$p$ sampling, which involves random variability in the generation process. We evaluate how consistent LLMs are with themselves through an inter-rater reliability (IRR) estimation. For each task, we interpret the three different LLM ratings as coming from three different annotators and we use the intra-class correlation coefficient (ICC), which is the most relevant one for our case study: unlike Cohen's and Fleiss's kappas \citep{cohen1960coefficient, fleiss1971measuring} or Krippendorff's alpha \citep{hayes2007answering}, which quantify IRR based on all-or-nothing agreement, the ICC incorporates the magnitude of the disagreement to compute its IRR estimate, with larger-magnitude disagreements resulting in lower ICC than smaller-magnitude disagreements \citep{hallgren2012computing}. We specifically use the ICC for \emph{average random raters} (ICC2k) \citep{vallat2018pingouin}; with the assumption that the \emph{random} aspect can approximate the random aspect of the generation.

ICC2k values for Eval-Prompt 1 for Beluga-13B, Mistral-7B and human ratings are displayed on \autoref{tab:icc1}. Comparing LLM consistency and human inter-rater agreement values should be done with caution: human raters may have subjective appreciations of the Likert scale despite guidelines, while LLM consistency depends mostly on parameters that dictate output variability, \textit{e.g.}\ temperature or top-$p$. 
That said, we reckon that it is still useful to display human IRR values as a baseline. We observe that LLMs have very high consistency overall for all criteria; the lowest value is Mistral-7B's ICC for Surprise (0.66), which is still fairly high. Confidence intervals are also smaller than for human ratings.

\subsubsection{Correlations with Human Annotations}
\label{ssub:correlations}

\begin{figure}[!h]
    \centering
    \includegraphics[width=\columnwidth]{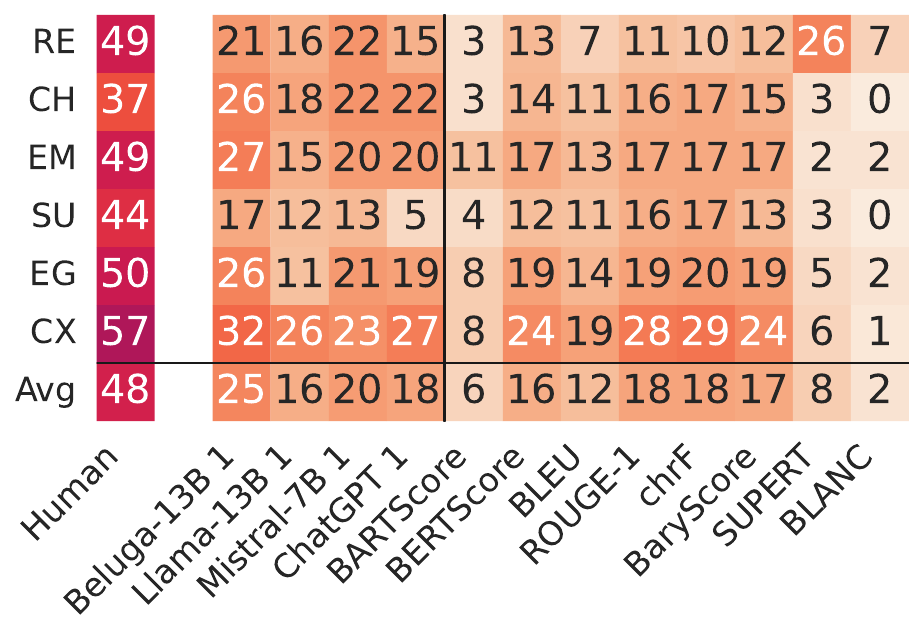}
    \caption{Overall absolute Kendall correlations between evaluation measures and human ratings. Higher is better. The black vertical line separates LLMs (left) and non-LLMs (right). Coefficient values are multiplied by 100 for readability; we will symbolize this with ``($\times$100)'' in the next figures.}
    \label{fig:story_level_kendall_mixed1_correlations}
\end{figure}

Here, we study the Kendall correlations between LLM and human ratings on corresponding criteria. For the ``Beluga-13B 1'' column in \autoref{fig:story_level_kendall_mixed1_correlations}, the first value is the correlation between Beluga-13B Relevance ratings and averaged human Relevance ratings for Eval-Prompt 1, then Coherence ratings, etc.

Assuming we want an automatic measure to perform as well as an individual human rater would, we need a baseline for comparison. Therefore, we also compute the average correlations between individual human ratings and 
average human ratings, which we compiled into the same figures for the sake of readability (the ``Human'' column). Since the individual human rating is included in the average human rating, both measures are not independent, so the column acts as an upper-bound.

\paragraph{Overall Correlations (\autoref{fig:story_level_kendall_mixed1_correlations}).}
LLM ratings generally correlate with human ratings similarly to automatic measures, if not better.
Overall, Beluga-13B is the best performer, achieving higher correlations (0.25 on average) than both other LLMs and automatic measures ($\leq$0.18). The better results (as compared to Llama-13B (0.16) and Mistral-7B (0.20)) suggest a positive influence of fine-tuning and model size respectively. The inferior performance of ChatGPT (0.18) is difficult to explain since OpenAI does not disclose the details of its architecture, its training process and, most importantly, its training data. Nonetheless, an important takeaway is that current source-available models can effectively compete with closed-source models: this is good news for NLP research, since observations made on closed-source models cannot easily be generalized.

\begin{figure}[!h]
    \centering
    \includegraphics[width=\columnwidth]{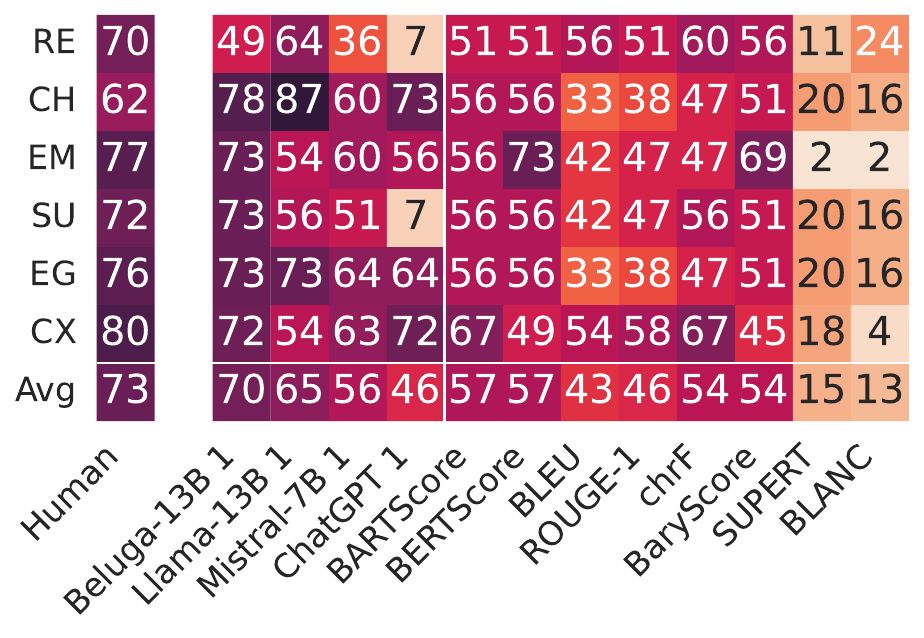}
    \caption{System-level absolute Kendall correlations ($\times$100) between evaluation measures and human ratings. Higher is better. The white vertical line separates LLMs (left) and non-LLMs (right).}
    \label{fig:system_level_kendall_mixed1_correlations}
\end{figure}

\paragraph{System-level Correlations (\autoref{fig:system_level_kendall_mixed1_correlations}).}
First, we observe that human baseline correlations are noticeably higher than non-LLM automatic measures: while human annotators tend to reach a consensus when ranking systems (averaging correlations of 0.73), non-LLM automatic measures are moderately to poorly correlated from human judgment (with values ranging from 0.13 to 0.57).

Meanwhile, Llama models display very high correlations, with Beluga-13B performing almost as well as human raters (0.70 vs 0.73). ChatGPT shows a somewhat erratic performance (correlations range from 0.07 to 0.73), which is overall comparable or inferior to Llama models. Also, LLMs generally outperform other automatic measures (0.70 for Beluga-13B compared to 0.57 for \texttt{BARTScore}).

The fact that correlations are sometimes higher than the baseline can be explained by the subjective nature of the task: human annotators may exhibit higher variability in their ratings than the stable LLMs.

\begin{table*}[!t]
\centering
\begin{tabular}{lcccc}
\toprule
Criterion & Eval-Prompt 1 & Eval-Prompt 2 & Eval-Prompt 3 & Eval-Prompt 4 \\
\midrule
Relevance & \result{0.88}{0.01} & \result{0.90}{0.01} & \result{0.85}{0.02} & \result{0.92}{0.01} \\
Coherence & \result{0.93}{0.01} & \result{0.94}{0.01} & \result{0.87}{0.01} & \result{0.93}{0.01} \\
Empathy & \result{0.88}{0.01} & \result{0.88}{0.01} & \result{0.83}{0.02} & \result{0.91}{0.01} \\
Surprise & \result{0.80}{0.02} & \result{0.79}{0.02} & \result{0.70}{0.03} & \result{0.85}{0.01} \\
Engagement & \result{0.91}{0.01} & \result{0.92}{0.01} & \result{0.79}{0.02} & \result{0.93}{0.01} \\
Complexity & \result{0.85}{0.01} & \result{0.86}{0.01} & \result{0.85}{0.01} & \result{0.89}{0.01} \\
\bottomrule
\end{tabular}
\caption{Intra-class coefficients type 2k for Beluga-13B ratings with 95\% confidence interval. Higher is better.}
\label{tab:icc2}
\end{table*}

\begin{table*}[!t]
\centering
\begin{tabular}{lcccc}
\toprule
LLM & Eval-Prompt 1 & Eval-Prompt 2 & Eval-Prompt 3 & Eval-Prompt 4 \\
\midrule
Beluga-13B & \result{3.48}{0.04} & \result{3.38}{0.03} & \result{3.06}{0.03} & \result{3.28}{0.04} \\
Llama-13B & \result{3.48}{0.03} & \result{3.52}{0.03} & \result{3.21}{0.02} & \result{2.82}{0.03} \\
Mistral-7B & \result{3.47}{0.03} & \result{3.51}{0.03} & \result{3.46}{0.03} & \result{3.28}{0.03} \\
ChatGPT* & \result{1.52}{0.03} & \result{1.47}{0.03} & \result{1.62}{0.02} & \result{1.60}{0.03} \\
\bottomrule
\end{tabular}
\caption{Average Likert ratings per LLM per Eval-Prompt. The asterisk signals the fact that ChatGPT was only asked to rate the original \texttt{HANNA} dataset without Llama-generated stories. Higher is better.}
\label{tab:average_ratings_per_llm}
\end{table*}

\paragraph{Statistical Testing.}

\begin{figure}[!h]
    \centering
    \includegraphics[width=\columnwidth]{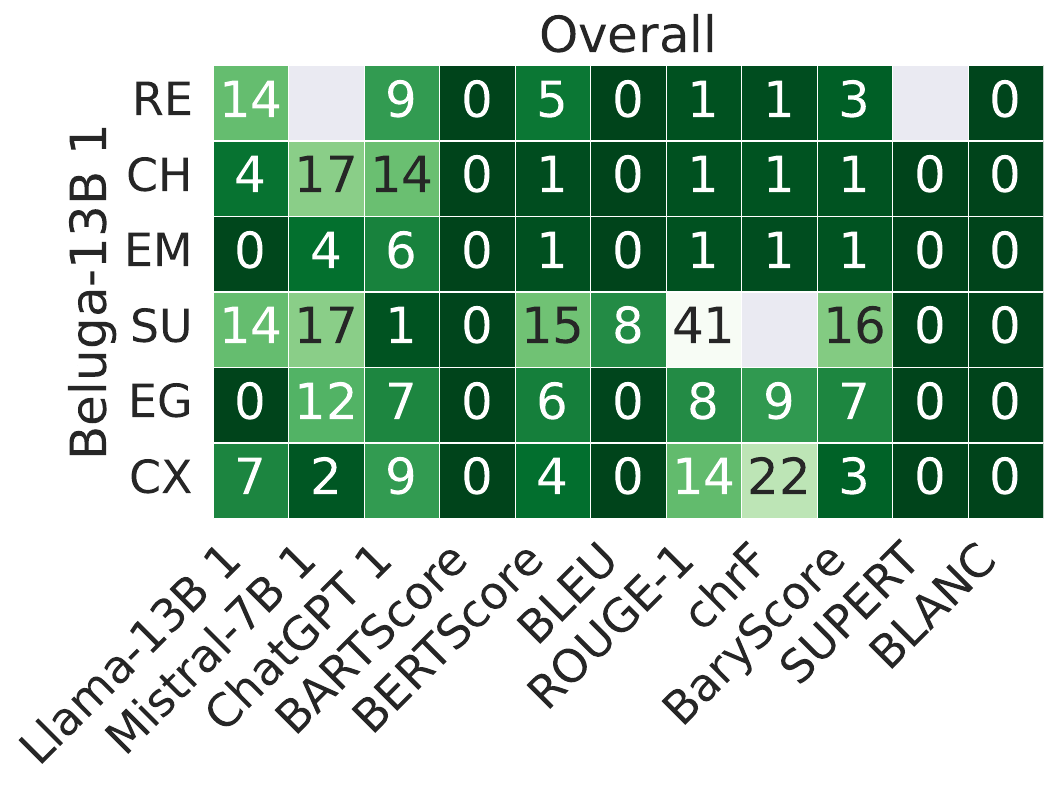}
    \includegraphics[width=\columnwidth]{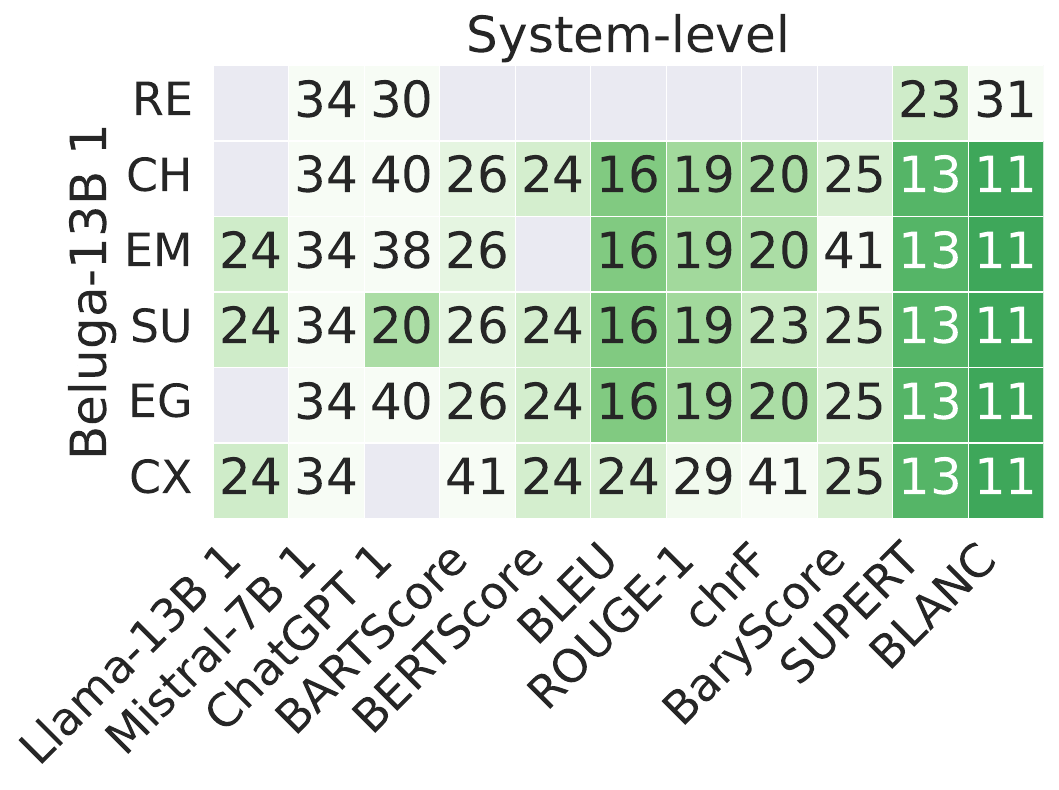}
    \caption{BH-adjusted $p$-values ($\times$100) of the Williams tests for overall and system-level Kendall correlations. Lower is better. ``0'' means $p<0.01$.}
    \label{fig:williams_beluga}
\end{figure}

\autoref{fig:williams_beluga} shows the BH-adjusted $p$-values of the Williams tests for the increase in correlations with a given criterion between Beluga-13B average Eval-Prompt~1 ratings (row) and other measures (column).

For overall correlations, there is strong statistical evidence that Beluga-13B correlates better with human judgment than many non-LLM automatic measures ($p$~<~0.01 for many tests). Evidence is more moderate to weak when comparing Beluga-13B and other LLMs. For instance, between Beluga-13B and ChatGPT, $p$-values lie between 0.01 and 0.14. While the performance of Beluga-13B still leaves a lot of room for improvement, it performs better than non-LLM automatic measures.

For system-level correlations, statistical evidence for better performance appears weaker: $p$~>~0.11 for all tests. However, one should keep in mind that the ratings (averaged over more than 1,000 stories) used to compute system-level correlations hold more information than the individual ratings of the overall correlations. Therefore, while statistical evidence is weaker, the averaged nature of the correlations and the significant numeric increases in correlations (0.70 for Beluga-13B vs 0.57 for \texttt{BARTScore}/\texttt{BERTScore}) suggest that Beluga-13B is more reliable at ordering systems compared to non-LLM measures.

\subsubsection{Takeaways}
First, LLMs show very high self-consistency. Overall correlations remain weak, although LLMs display marginal improvements over non-LLM automatic measures, backed with strong statistical evidence. At the system-level, LLM correlations with human judgment are high, but statistical evidence is weaker. In conclusion, while LLMs still cannot be relied upon to evaluate a single story, they appear more reliable than non-LLM automatic measures for comparing different models and selecting the best one. 

\subsection{ASE2: Influence of the Eval-Prompt}
\label{sub:ase2_analysis}
In this section, we discuss the influence of the Eval-Prompt on the consistency and distribution of the generated LLM ratings.

\subsubsection{Influence on Consistency}
\label{sub:influence_icc}
Here, we analyse the influence of the Eval-Prompt on LLM consistency. ICC2k values for Beluga-13B ratings w.r.t.\ the different Eval-Prompts are shown on \autoref{tab:icc2} (other LLMs display similar behavior). The influence of Eval-Prompts appears limited: providing guidelines (Eval-Prompt 3) tends to decrease self-consistency for all criteria except Complexity with a discernible effect (as shown by the confidence intervals), but ICC values remain very high. LLMs are therefore remarkably consistent in their grading, no matter the Eval-Prompt.

\subsubsection{Influence on Ratings}
\label{ssub:influence_ratings}

We show the average Likert ratings per LLM per Eval-Prompt on \autoref{tab:average_ratings_per_llm}. Compared to Eval-Prompt~1, Eval-Prompt 2 seems to have limited influence on the ratings for all models, often leading to overlapping confidence intervals. Eval-Prompt 3 causes a statistically discernible decrease in ratings for Beluga-13B and Llama-13B, and a discernible increase for ChatGPT. Eval-Prompt 4 has a similar effect, with the decrease also observable with Mistral-7B. The significantly lower ratings of ChatGPT partly stem from the fact that it was not asked to rate the new Llama-generated stories, which were generally highly-rated.

Overall, it seems that more detailed Eval-Prompts (3 and 4) tend to decrease the ratings for Llama-models while having an opposite effect for ChatGPT. We tried to separate ratings per generative model or per criterion but were unable to identify a more specific pattern: we therefore chose to show only the aggregated results for the sake of clarity.

\subsubsection{Influence on Correlations}
\label{ssub:influence_correlations}
Here we analyze the influence of Eval-Prompts on correlations between LLM ratings and human ratings.

\begin{figure}[!h]
    \centering
    \includegraphics[width=\columnwidth]{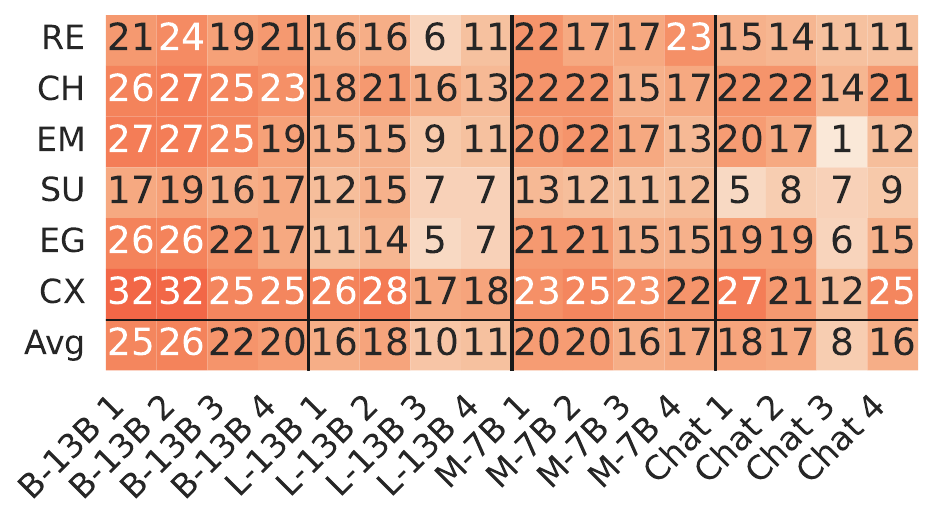}
    \caption{Overall absolute Kendall correlations ($\times$100) between LLMs and human ratings for different Eval-Prompts. Higher is better. B-13B = Beluga-13B, L-13B = Llama-13B, M-7B = Mistral-7B and Chat = ChatGPT.}
    \label{fig:story_level_kendall_mixed2_correlations}
\end{figure}

\paragraph{Overall Correlations (\autoref{fig:story_level_kendall_mixed2_correlations}).}
Eval-Prompt 2 overall correlations are very close to Eval-Prompt~1 correlations for all models: simply asking for an explanation has limited influence on correlations. Eval-Prompt 3 tends to decrease correlations for all models: providing guidelines makes the model less accurate, counter-intuitively. Eval-Prompt 4 (providing a human story for reference) has a similar effect.

\begin{figure}[!h]
    \centering
    \includegraphics[width=\columnwidth]{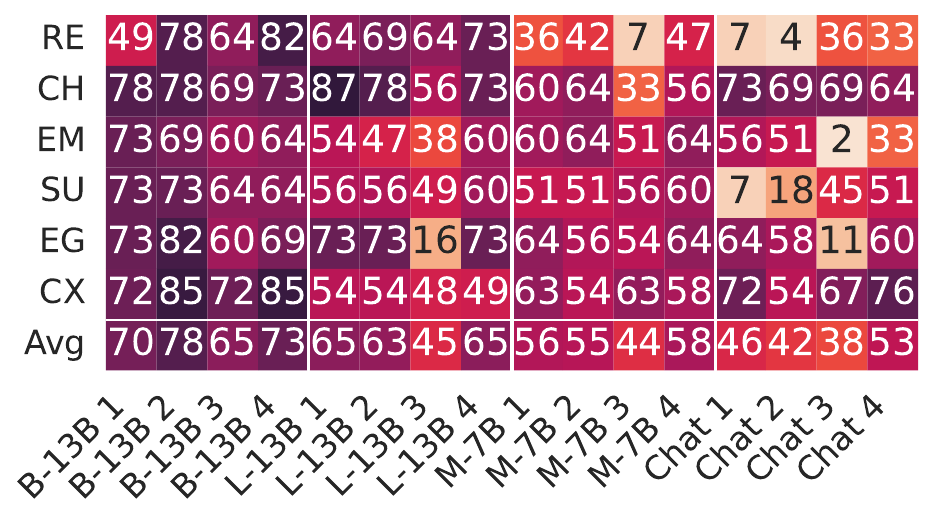}
    \caption{System-level absolute Kendall correlations ($\times$100) between LLMs and human ratings for different Eval-Prompts. Higher is better. B-13B = Beluga-13B, L-13B = Llama-13B, M-7B = Mistral-7B and Chat = ChatGPT.}
    \label{fig:system_level_kendall_mixed2_correlations}
\end{figure}

\paragraph{System-level Correlations (\autoref{fig:system_level_kendall_mixed2_correlations}).}

Eval-Prompt~2 has limited effect on correlations again, except for Beluga-13B for whom it seems to increase correlations. Eval-Prompt 3 decreases correlations, with a marked effect in Llama-13B. Finally, Eval-Prompt~4 seems to cause a small increase in correlations, contrary to its decreasing effect on overall correlations.

\subsubsection{Takeaways}
First, regardless of Eval-Prompt complexity, LLMs behave consistently when prompted multiple times. Asking for an explanation (Eval-Prompt 2) has negligible effect on ratings, while more complex Eval-Prompts (3 - providing guidelines and 4 - providing a reference human story) have a more discernible influence (positive or negative). As for correlations with human ratings, providing guidelines (Eval-Prompt 3) consistently seems to lower correlations, whereas providing a human story for reference (Eval-Prompt 4) has opposite effects for overall or system-level correlations.

\subsection{ASE3: Explainability of Ratings}
\label{sub:ase3_analysis}

In this section, we analyze to what extent the explanations provided by LLMs are consistent w.r.t.\ their ratings, \textit{i.e.},\ whether they differ from criterion to criterion, whether they are semantically relevant and, for Eval-Prompt 3, whether they are compliant with the provided guidelines. We will focus on Beluga-13B since it had the best correlations with human judgment, as shown in \autoref{sub:ase1_analysis}.

\subsubsection{Visualization of Explanation Embeddings}

\begin{figure}[h!]
\centering
\includegraphics[width=\columnwidth]{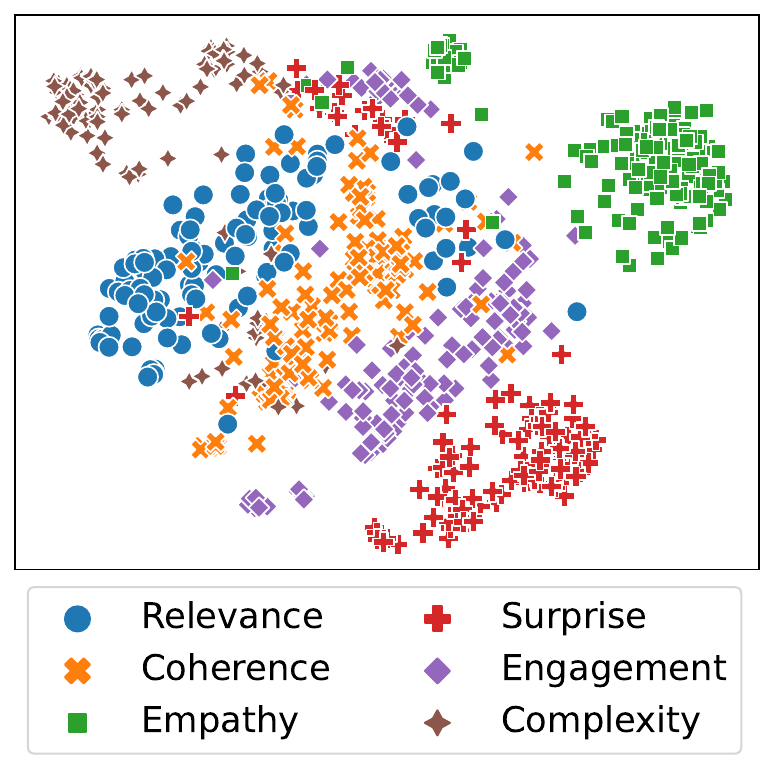}
\caption{UMAP projection of Beluga-13B explanations.}
\label{fig:umap_explanation}
\end{figure}

First, we want to ascertain whether Beluga-13B provides different explanations for each of the human criteria. We gather the explanations provided by Beluga-13B on human stories for each criterion and use the \textbf{SentenceTransformers} library \citep{reimers-2019-sentence-bert} to compute their corresponding embeddings.
We then use a 2D UMAP projection \citep{mcinnes2018umap} (with parameters $\textrm{n\_neighbors}=300$ and $\textrm{metric}=\textrm{euclidean}$) to visualize how the embeddings are distributed. \autoref{fig:umap_explanation} shows the visualization of the UMAP projection: Beluga's explanations are overall well-separated w.r.t.\ their corresponding criteria. 

\subsubsection{Keyword Analysis}

Since Beluga's explanations seem to vary from one criterion to another, we evaluate whether they make sense from a semantic point of view
. We use the YAKE! keyword extractor, which significantly outperforms other state-of-the-art methods  \citep{campos2020yake}: we show selected 3-gram keywords from the top-30 per criterion on \autoref{tab:explanation_keywords}. The results are consistent with \autoref{fig:umap_explanation}: keywords are overall different for each criterion. We can also see here that they are semantically relevant.

\begin{table}[!h]
\centering
\begin{tabular}{cp{0.8\linewidth}}
\toprule
Crit. & Keywords\\
\midrule
RE & story, prompt, roughly matches, target, weak relationship, connection, weak\\
\midrule
CH & story, coherence, make sense, difficult to understand, clear narrative structure\\
\midrule
EM & empathy, emotions, understand the characters, depth, emotional connection\\
\midrule
SU & story, surprise, ending, predictable, rate, unexpected, twist, completely obvious\\
\midrule
EG & story, mildly interesting, engagement, difficult, found, characters, fully engage\\
\midrule
CX &  story, characters, intricate plot, difficult to understand, straightforward, depth\\
\bottomrule
\end{tabular}
\caption{Selected keywords from Beluga-13B explanations w.r.t.\ a specific criterion.}
\label{tab:explanation_keywords}
\end{table}

\begin{table}[!t]
\centering
\begin{tabular}{lcc}
\toprule
Error Type & Rate & AC1\\
\midrule
Poor Syntax & 0.02 & \icc{0.93}{0.97}{1.00}\\
Incoherence & 0.11 & \icc{0.73}{0.81}{0.89}\\
Wrong Guideline & 0.13 & \icc{0.85}{0.90}{0.96}\\
Superfluous Text & 0.20 & \icc{0.55}{0.66}{0.78}\\
Unsubstantiated Claims & 0.31 & \icc{0.47}{0.60}{0.74}\\
\bottomrule
\end{tabular}
\caption{Error rates of Beluga-13B Eval-Prompt 3 on a sample of 100 explanations. Lower is better.}
\label{tab:error_rates}
\end{table}

\subsubsection{User Study on LLM Explanations}
\label{ssub:user_study_results}

We display the results of our user study (designed in \autoref{ssub:user_study}) in \autoref{tab:error_rates}. We also display the IRR, which we computed using Gwet's agreement coefficient 1 (AC1) \citep{gwet2008computing, fergadis2022irrcac}. Gwet's AC1 is known to perform well for IRR estimation on binary classification tasks such as our user study: it was designed to be more stable and less affected by prevalence and marginal probability than Cohen's kappa, and this was confirmed by practical experiments \citep{wongpakaran2013comparison}.

We can see that Beluga-13B produces near-impeccable syntax, at least according to annotators (2\% of ``Poor Syntax''). It also does a good job at producing coherent text (11\% of ``Incoherence''), and mostly understands the guidelines (13\% of ``Wrong Guideline''). However, it tends to repeat itself somewhat (20\% of ``Superfluous Text'') and, most notably, tends not to substantiate its claims with direct references to the story (31\% of ``Unsubstantiated Claims''). Overall, annotators tend to agree with one another, as showed by the high values of Gwet's AC1.

The substantial rate of ``Unsubstantiated Claims'' and the fact that \textbf{40\% of all Eval-Prompt 3 ratings are not supported by an explanation}---despite the Eval-Prompt explicitly asking for it---beg the question of whether Beluga-13B truly understands the given task. We discuss this question further in \autoref{sec:discussion}.

\paragraph{Takeaways.}
LLM explanations seem to be specific to each considered human evaluation criterion; however, a finer analysis with a user study reveals that LLMs often struggle with following guidelines and substantiating their explanations.

\begin{table*}[!t]
\small
\centering
\begin{tabular}{lccccccc}
\toprule
Model &            RE &            CH &              EM &             SU &           EG &           CX &              Average \\
\midrule
Human             &  \result{3.37}{0.12} &  \result{3.55}{0.11} &  \result{3.42}{0.11} &  \result{3.11}{0.13} &  \result{3.58}{0.10} &  \result{3.48}{0.10} &  \result{3.42}{0.06} \\
\midrule
Platypus2-70B     &  \result{4.09}{0.05} &  \result{4.31}{0.05} &  \result{3.92}{0.06} &  \result{\textbf{3.69}}{0.07} &  \result{4.19}{0.05} &  \result{3.88}{0.05} &  \result{4.01}{0.03} \\
Llama-30B &  \result{\textbf{4.19}}{0.05} &  \result{\textbf{4.38}}{0.04} &  \result{\textbf{4.04}}{0.06} &  \result{\textbf{3.63}}{0.09} &  \result{\textbf{4.31}}{0.05} &  \result{\textbf{3.98}}{0.05} &  \result{\textbf{4.08}}{0.03} \\
Beluga-13B        &  \result{4.06}{0.08} &  \result{4.10}{0.06} &  \result{3.75}{0.08} &  \result{3.54}{0.08} &  \result{3.90}{0.08} &  \result{3.69}{0.07} &  \result{3.84}{0.05} \\
Mistral-7B    &  \result{4.12}{0.05} &  \result{4.25}{0.05} &  \result{3.86}{0.06} &  \result{3.56}{0.08} &  \result{4.11}{0.05} &  \result{3.82}{0.04} &  \result{3.95}{0.03} \\
Llama-7B      &  \result{4.07}{0.06} &  \result{4.24}{0.05} &  \result{3.90}{0.06} &  \result{3.58}{0.06} &  \result{4.09}{0.05} &  \result{3.79}{0.05} &  \result{3.95}{0.03} \\
GPT-2             &  \result{2.57}{0.13} &  \result{2.36}{0.11} &  \result{2.72}{0.11} &  \result{2.59}{0.14} &  \result{2.67}{0.12} &  \result{2.89}{0.12} &  \result{2.63}{0.07} \\
HINT              &  \result{1.57}{0.10} &  \result{1.31}{0.07} &  \result{1.59}{0.10} &  \result{1.49}{0.10} &  \result{1.58}{0.09} &  \result{1.43}{0.08} &  \result{1.49}{0.06} \\
\bottomrule
\end{tabular}
\caption{Average Beluga-13B ratings for Eval-Prompt 1 with 95\% confidence interval. Higher is better.}
\label{tab:average_beluga_ratings}
\end{table*}

\begin{table*}[!ht]
\small
\centering
\begin{tabular}{lccccccc}
\toprule
Model &            RE &            CH &              EM &             SU &           EG &           CX &              Average \\
\midrule
Human             &  \result{3.48}{0.11} &  \result{3.50}{0.10} &  \result{3.69}{0.08} &  \result{3.24}{0.11} &  \result{3.42}{0.10} &  \result{3.45}{0.07} &  \result{3.46}{0.05} \\
\midrule
Platypus2-70B     &  \result{\textbf{4.26}}{0.08} &  \result{\textbf{4.31}}{0.08} &  \result{\textbf{4.05}}{0.07} &  \result{\textbf{3.46}}{0.10} &  \result{\textbf{3.94}}{0.06} &  \result{3.55}{0.07} &  \result{\textbf{3.93}}{0.03} \\
Llama-30B &  \result{4.15}{0.10} &  \result{\textbf{4.29}}{0.07} &  \result{\textbf{4.02}}{0.07} &  \result{\textbf{3.46}}{0.09} &  \result{\textbf{3.94}}{0.06} &  \result{\textbf{3.65}}{0.07} &  \result{\textbf{3.92}}{0.03} \\
Beluga-13B        &  \result{4.07}{0.09} &  \result{4.14}{0.07} &  \result{3.98}{0.07} &  \result{\textbf{3.50}}{0.09} &  \result{3.74}{0.08} &  \result{3.59}{0.07} &  \result{3.84}{0.03} \\
Mistral-7B    &  \result{4.15}{0.10} &  \result{4.22}{0.08} &  \result{\textbf{4.02}}{0.07} &  \result{\textbf{3.51}}{0.11} &  \result{\textbf{3.94}}{0.07} &  \result{\textbf{3.67}}{0.07} &  \result{\textbf{3.92}}{0.04} \\
Llama-7B      &  \result{4.13}{0.10} &  \result{4.14}{0.09} &  \result{3.90}{0.08} &  \result{\textbf{3.48}}{0.09} &  \result{3.78}{0.08} &  \result{3.56}{0.08} &  \result{3.83}{0.05} \\
GPT-2             &  \result{2.40}{0.10} &  \result{2.37}{0.09} &  \result{2.74}{0.10} &  \result{2.85}{0.11} &  \result{2.60}{0.09} &  \result{2.88}{0.09} &  \result{2.64}{0.05} \\
HINT              &  \result{2.12}{0.11} &  \result{2.13}{0.08} &  \result{2.23}{0.10} &  \result{2.28}{0.11} &  \result{2.05}{0.08} &  \result{2.05}{0.09} &  \result{2.15}{0.06} \\
\bottomrule
\end{tabular}
\caption{Average Mistral-7B ratings for Eval-Prompt 1 with 95\% confidence interval. Higher is better.}
\label{tab:average_mistral_ratings}
\end{table*} 

\subsection{ASG1: LLM Performance in ASG}
\label{sub:ASG1_analysis}
In this section, we discuss the performance of LLMs at the ASG task compared to human and previous models' performance, as we expanded the \texttt{HANNA} dataset with stories generated from more recent models. Since Beluga-13B and Mistral-7B display very high system-level correlations with human ratings (see \autoref{fig:system_level_kendall_mixed1_correlations}), we use their ratings as proxy for human ratings. \autoref{tab:average_beluga_ratings} and \autoref{tab:average_mistral_ratings} show the average Beluga-13B and Mistral-7B ratings for Eval-Prompt 1 per model per criterion for a few \texttt{HANNA} models (GPT-2, HINT) and the Llama models.

We observe that LLMs perform remarkably well, getting higher ratings than older models (GPT-2) and even human stories. Beluga-13B and Mistral-7B both seem to prefer the outputs from larger LLMs (Platypus2-70B, Llama-30B) to their own outputs, suggesting that the LLM grading process cannot be explained simply by a proxy for perplexity. Interestingly, in both tables, Mistral-7B gets slightly higher ratings than Beluga, with some differences being statistically discernible, which could be explained by differences in fine-tuning data.

\paragraph{Takeaways.} Larger models (Platypus2-70B, Llama-30B) exhibit the best ASG performance, with LLM ratings at least equal to those of human stories. However, our setting involves short stories of between 500 and 1,000 words; generating longer stories may prove more difficult since maintaining large-scale coherence may become an issue.

\subsection{ASG2: Influence of Pretraining Data on ASG Performance}
\label{sub:ASG2_analysis}

In this section, we verify whether the LLM pretraining data contains the WritingPrompts dataset to check for model contamination, as advised by \citet{magar-schwartz-2022-data}, and to what extent ASG performance is related with data exploitation, \textit{e.g.}\ through reproduction of training examples.

We use the \minkprob\ detection method \citep{shi2023detecting} which is based on the hypothesis that unseen data will contain more outlier words with low probability than seen data. Furthermore, it does not require additional training. Given a sentence and an LLM's probability distribution of the next token, \minkprob\ selects the top-$k$\% of tokens with the highest negative log-likelihood and computes their average log-likelihood. We can then detect if the sentence was included in pretraining data by thresholding this average. We follow \citet{shi2023detecting} and use $k=20$ for our two experiments.

\begin{table}[h]
\centering
\begin{tabular}{lr}
\toprule
Model & Contamination (\%) \\
\midrule
Platypus2-70B & 0.80 \\
Llama-30B & 1.80 \\
Beluga-13B & 4.40\\
Mistral-7B & 2.50 \\
Llama-7B & 10.10\\
\bottomrule
\end{tabular}
\caption{Predicted contamination rates of the WritingPrompts sample.}
\label{tab:wp_contamination}
\end{table}

\paragraph{Model Contamination.} 
We sample 1,000 stories from the WritingPrompts dataset \citep{fan2018hierarchical}, from which the \texttt{HANNA} human stories come. \autoref{tab:wp_contamination} shows the predicted contamination rates of the WritingPrompts sample. Since they are very low, this strongly suggests that the WritingPrompts sample was not included in the pretraining data of the evaluated models. We can reasonably surmise that the same applies to the whole WritingPrompts dataset.

\paragraph{Data Reproduction.}
We use the BooksMIA dataset \citep{shi2023detecting}, which contains 9,870 samples of books labeled 0 if included in the Books3 dataset (commonly used for pretraining LLMs) or 1 if released in or after January 2023. Since the BooksMIA data is labeled, we compute the area under the ROC curve (AUC) obtained with \minkprob\ thresholding. Results are shown on \autoref{tab:book_auc}.

\begin{table}[h]
\centering
\begin{tabular}{lc}
\toprule
Model & AUC (\%) \\
\midrule
Platypus2-70B & 92.1 \\
Llama-30B & 81.3 \\
Beluga-13B & 70.1\\
Mistral-7B & 51.2 \\
Llama-7B & 55.1 \\
\bottomrule
\end{tabular}
\caption{AUC detection score on the BooksMIA dataset.}
\label{tab:book_auc}
\end{table}

We observe that the AUC detection score is higher for larger models, \textit{i.e.}, it is easier to detect if a book was in the pretraining data of a larger LLM. The definition of the \minkprob\ measure also means that larger LLMs tend to produce text that is more similar to their pretraining data, such as fiction books, which could help explain their better ASE ratings.

\paragraph{Takeaways.} The better performance of larger LLMs for ASG may be partially explained by their tendency to generate text that is more similar to their pretraining data, \textit{e.g.}\ existing novels.

\section{Discussion on LLM performance}
\label{sec:discussion}
Our work is part of the ongoing research on the general ability of LLMs for understanding and thinking.

\citet{mahowald2023dissociating} distinguish formal (the statistical features of language) and functional linguistic competence (the ability to use language in the world) and show that LLMs are very successful on formal linguistic tasks but struggle at functional linguistic tasks. \citet{bubeck2023sparks} argue that LLMs do display impressive performance at a wide variety of tasks but lack ``slow thinking'' capabilities, referring to the System 1--System 2 dichotomy introduced by \citet{kahneman2011thinking}.

Thus, the high performance of LLMs at ASE should be interpreted with caution: we hypothesize that the ``rating'' part of our story evaluation experiments could be linked to formal linguistic competence and the fast, automatic System 1, while the ``explanation'' part would correspond to functional linguistic competence and the slow, conscious System 2.

This analogy would explain the good correlations of LLM ratings with human ratings: the internal criterion of LLMs for story evaluation may be formal quality (vocabulary, syntax, grammar), regardless of the criterion mentioned in the Eval-Prompt. Indeed, the six criteria from \citet{chhun-etal-2022-human} are mostly orthogonal but not completely independent: their correlation with one another may be related to the general ``System 1'' tendency of human raters to favour stories that display better formal qualities. In that sense, LLMs may reflect a human bias towards easy, intuitive thinking. By contrast, the less convincing performance of LLMs at explaining their ratings may highlight their weaker System 2 capabilities as argued by \citet{mahowald2023dissociating} and \citet{bubeck2023sparks}.


\section{Conclusions}
\label{sec:conclusions}

\subsection{Practical Takeaways}
\label{sub:practical_takeaways}

\begin{enumerate}[wide, labelindent=0pt, noitemsep]
    \item \textbf{Used with prompts based on specific criteria, LLMs are currently the best proxy for human evaluation of story generation (\autoref{ssub:correlations}).} In particular, LLMs display very high system-level correlations with human judgment;
    \item \textbf{LLMs are remarkably self-consistent (\autoref{ssub:icc1}),} exhibiting very high intra-class coefficient values;
    \item \textbf{LLMs understand the ASE task only partially (\autoref{ssub:user_study_results}):} they struggle to explain their answers with substantiated claims;
    \item \textbf{For ASE, providing detailed guidelines (Eval-Prompt 3) did not lead to improved correlations with human ratings (\autoref{ssub:influence_correlations}).} Providing a human story for reference (Eval-Prompt 4) yields mixed results;
    \item \textbf{LLM stories have at least equal ASE ratings to human stories (\autoref{sub:ASG1_analysis}),} with larger LLMs exhibiting the best performance;
    \item \textbf{Pretraining data helps explain LLM performance at ASG (\autoref{sub:ASG2_analysis}):} the higher ratings of larger LLMs may be due to their ability to produce output similar to existing books.
\end{enumerate}

\subsection{Limitations and Future Directions}
The ASE task is a very subjective one: LLM performance at ASE and ASG must be seen as a reflection of \textit{average} preferences and may therefore include biases, \textit{e.g.}\ from their pretraining data.

Furthermore, we performed most of our experiments in a zero-shot setting without further training; it would be interesting to compare our results with future work involving fine-tuning or reinforcement learning with human feedback on data specific to ASE.

Also, we did not conduct our experiments with LLMs that were optimized for long inputs and outputs, such as GPT-4.

Finally, we mainly used source-available LLama models and found that they performed at least as well as ChatGPT, a proprietary model. We encourage the NLP community to favor the use of such models, as the growing presence of closed models hinders research transparency and reproductibility.



\subsection*{Acknowledgments}
This work was performed using HPC resources from GENCI-IDRIS (Grant 2022-AD011013105R1) and was partially funded by the grants ANR-20-CHIA-0012-01 (``NoRDF'') and ANR-23-CE23-0033-01 (``SINNet'').

We would also like to convey our appreciation to TACL Action Editor Ehud Reiter, as well as to our anonymous reviewers, for their valuable feedback.

\bibliography{tacl}
\bibliographystyle{acl_natbib}

\end{document}